\documentclass[10pt,twocolumn,letterpaper]{article}

\usepackage{cvpr}              

\definecolor{cvprblue}{rgb}{0.21,0.49,0.74}
\usepackage[pagebackref,breaklinks,colorlinks,allcolors=cvprblue]{hyperref}

\def\paperID{*****} 
\def\confName{CVPR}
\def\confYear{2025}

\title{Elevation Aware 2D/3D Co-simulation Framework for Large-scale Traffic Flow and High-fidelity Vehicle Dynamics}

\author{
Chandra Raskoti \quad
Weizi Li \\
University of Tennessee, Knoxville, USA\\
{\tt\small craskoti@vols.utk.edu, weizili@utk.edu}
}

\begin{document}
\maketitle
\begin{abstract}

Reliable testing of autonomous driving systems requires simulation environments that combine large-scale traffic modeling with realistic 3D perception and terrain. Existing tools rarely capture real-world elevation, limiting their usefulness in cities with complex topography. This paper presents an automated, elevation-aware co-simulation framework that integrates SUMO with CARLA using a pipeline that fuses OpenStreetMap road networks and USGS elevation data into physically consistent 3D environments. The system generates smooth elevation profiles, validates geometric accuracy, and enables synchronized 2D–3D simulation across platforms. Demonstrations on multiple regions of San Francisco show the framework’s scalability and ability to reproduce steep and irregular terrain. The result is a practical foundation for high-fidelity autonomous vehicle testing in realistic, elevation-rich urban settings.
\end{abstract}    
\section{Introduction}
\label{sec:intro}

The advent of Connected and Automated Vehicles (CAVs) represents a paradigm shift in the transportation industry, offering substantial potential for improving traffic safety and quality~\cite{Villarreal2024AutoJoin,Lin2022Attention,Poudel2022Micro,Shen2021Corruption,Shen2022IRL,Li2019ADAPS}. The development and validation of these complex systems require extensive testing to ensure their reliable operation in a myriad of real-world scenarios~\cite{cantas2023customizedcosimulationenvironmentautonomous}. However, on-road testing is a costly, time-consuming, and potentially hazardous endeavor, particularly when exploring a ``long tail'' of rare or critical events~\cite{cantas2023customizedcosimulationenvironmentautonomous}. Consequently, virtual simulation has emerged as the most viable and effective method for the development, training, and validation of autonomous driving systems (ADS)~\cite{cantas2023customizedcosimulationenvironmentautonomous}. It provides a safe, controllable, and cost-effective environment where engineers can rapidly iterate on algorithms and test them against diverse and potentially dangerous scenarios that are difficult or impossible to replicate in the real world~\cite{cantas2023customizedcosimulationenvironmentautonomous}.
A significant challenge in virtual testing is achieving a balance between the fidelity of the simulation and its scalability. High-fidelity simulators, such as CARLA, are widely used for end-to-end testing of ADS because they provide photorealistic 3D environments, detailed vehicle dynamics, and a rich suite of realistic sensors, including LiDAR, radar, and cameras~\cite{cantas2023customizedcosimulationenvironmentautonomous}. These simulators are essential for developing and testing perception-based control systems and other advanced functions that rely on a realistic representation of the vehicle's surroundings~\cite{cantas2023customizedcosimulationenvironmentautonomous}. However, the computational demands of rendering and simulating vehicle dynamics for a large number of vehicles make these platforms impractical for large-scale urban traffic scenarios~\cite{cantas2023customizedcosimulationenvironmentautonomous}. In contrast, microscopic traffic simulators like Simulation of Urban Mobility (SUMO) are highly efficient at modeling the behavior of thousands of individual vehicles across vast road networks but typically lack the high-fidelity 3D environments, realistic sensor models, and detailed vehicle dynamics needed for AV research~\cite{cantas2023customizedcosimulationenvironmentautonomous}. The limitations of a single simulation tool necessitate a co-simulation approach that combines their respective strengths~\cite{cantas2023customizedcosimulationenvironmentautonomous}.

A critical hurdle for co-simulation frameworks is the gap between the simulated world and reality, often referred to as the ``domain gap''~\cite{12-06-01-0007}. Creating valid traffic scenarios requires a virtual environment that is a faithful representation of a real-world location, a concept known as a ``digital twin''~\cite{cantas2023customizedcosimulationenvironmentautonomous}. Manually creating these environments is a laborious and error-prone process~\cite{cantas2023customizedcosimulationenvironmentautonomous}. Moreover, many existing procedural generation methods focus on idealized, flat urban layouts, which fail to capture the complexities of real-world topography~\cite{cantas2023customizedcosimulationenvironmentautonomous}. This is particularly problematic for cities with significant elevation changes, such as San Francisco, where elevated road structures, steep gradients, and sharp curves are common. These complex geometries present unique and difficult challenges for autonomous vehicles, impacting everything from perception—where vehicle-mounted sensors may be occluded by crests to control, which must manage vertical acceleration and dynamic stability on steep slopes~\cite{s24020452}.
To address these challenges, this paper presents an integrated and automated workflow for generating high-fidelity, elevation-aware 3D environments from open-source Geographic Information System (GIS) data. This methodology leverages a multi-stage pipeline to transform readily available 2D road networks and digital elevation data into a geometrically and topologically consistent 3D scene that can be seamlessly integrated into a CARLA-SUMO co-simulation framework. The resulting platform provides a robust and realistic testing environment that is essential for advancing research into autonomous vehicle performance in challenging urban topographies.

This study makes the following key contributions:
\begin{itemize}
    \item A semi-automated, and modular pipeline for constructing high-fidelity, elevation-aware 3D road networks. This pipeline synthesizes data from OpenStreetMap (OSM) and the U.S. Geological Survey (USGS), utilizes the procedural generation capabilities of RoadRunner, and culminates in a CARLA-compatible environment that accurately represents real-world topographies.
    \item A demonstration of a scalable CARLA-SUMO co-simulation running on these complex, elevated urban road networks, showcasing the framework's utility for advanced AV research and perception system validation in challenging conditions. The framework is designed to run the ego vehicle's control and sensor integration within CARLA, while SUMO manages the large-scale background traffic.
    \item An evaluation of the framework's capabilities for integrating realistic sensor data, particularly LiDAR point clouds, in a dynamic, elevation-aware environment. The analysis highlights how such a platform can be used to study specific perception challenges, such as sensor occlusion on crests and the impact of road geometry on object detection.
\end{itemize}

\section{Related Work}
\label{sec:related_work}

This section reviews the relevant literature across four key areas: traffic simulation frameworks, co-simulation approaches, 3D road network generation, and elevation-aware simulation methodologies.

\subsection{Traffic Simulation Frameworks}

Traffic simulation has evolved significantly from early macroscopic models to sophisticated microscopic and mesoscopic simulation frameworks~\cite{Guo2024Simulation,Chao2020Survey,Li2017CityFlowRecon,Wilkie2015Virtual}. SUMO~\cite{lopez2018microscopic} has emerged as one of the most widely used open-source microscopic traffic simulators, offering detailed vehicle behavior modeling, traffic signal control, and multi-modal transportation simulation. However, SUMO operates primarily in 2D space and lacks the visual fidelity required for computer vision applications.

CARLA~\cite{dosovitskiy2017carla} represents a significant advancement in autonomous vehicle simulation, providing photorealistic 3D environments with comprehensive sensor simulation capabilities including cameras, LiDAR, radar, and GPS. While CARLA excels in visual simulation and sensor modeling, its traffic simulation capabilities are more limited compared to dedicated traffic simulators like SUMO.

Other notable traffic simulation platforms include VISSIM, AIMSUN, and MATSim, each with specific strengths in different simulation aspects. However, these platforms typically focus on either 2D traffic flow modeling or 3D visualization, rarely combining both effectively while incorporating real-world topographical data.

\subsection{Co-Simulation Approaches}

Recognizing the complementary strengths of different simulation platforms, researchers have developed various co-simulation frameworks that integrate multiple simulators. Li et al.~\cite{li2021novel} presented a framework combining SUMO and CARLA for autonomous vehicle testing, demonstrating improved realism in traffic scenarios. Their approach focuses on synchronizing traffic flow between the two platforms but does not address topographical considerations.

Azfar and Ke~\cite{azfar2022traffic} developed a co-simulation framework enhanced with infrastructure camera sensing and reinforcement learning for adaptive traffic signal control. While their work advances the state of traffic management, it operates on relatively flat road networks without complex elevation variations.

Zhang et al.~\cite{zhang2023virtual} introduced a digital twin platform integrating CARLA, SUMO, and NVIDIA PhysX for mixed autonomous traffic safety analysis. Their comprehensive approach includes real-world data integration but requires significant manual effort for 3D environment creation.

The OpenCDA framework~\cite{xu2021opencda} provides an open cooperative driving automation platform integrated with co-simulation capabilities. However, like most existing frameworks, it does not systematically address the challenges of complex urban topographies with significant elevation variations.

\subsection{3D Road Network Generation}

The generation of accurate 3D road networks from real-world data has been an active area of research in computer graphics and GIS communities. Traditional approaches rely heavily on manual modeling or semi-automated processes that require significant human intervention.

Recent work has explored the integration of OpenStreetMap data with elevation information for 3D visualization. However, most existing methods focus on basic visualization rather than creating simulation-ready environments with proper lane-level detail and traffic infrastructure.

RoadRunner, developed by MathWorks, provides advanced capabilities for creating 3D road networks, but typically requires manual design or significant preprocessing of input data. While it offers excellent tools for fine-tuning road networks, the automation of the entire pipeline from raw GIS data to simulation-ready environments remains a challenge.

\subsection{Elevation-Aware Simulation}
Traffic data have been used extensively in traffic state forecasting and vehicle trajectory prediction~\cite{Poudel2025Urban,Raskoti2025MIAT,Lin2022GCGRNN,Lin2019BikeTRB,Poudel2021Attack,Lin2019Compress,Li2018CityEstIET,Li2017CitySparseITSM}.  
The importance of elevation data in traffic simulation has been recognized in several domains. Energy consumption models for electric vehicles specifically account for elevation changes, as they significantly impact vehicle range and performance. However, these models typically operate with simplified elevation profiles rather than full 3D topographical representations.

In the context of autonomous vehicle simulation, elevation information affects sensor performance, particularly for LiDAR and camera systems where line-of-sight and field-of-view considerations are critical. Despite this importance, most existing autonomous vehicle simulation frameworks operate on relatively flat synthetic environments.

Some recent work has begun to explore the integration of Digital Elevation Models (DEMs) with road network data, but these efforts are typically limited to specific geographic regions or require extensive manual processing.

\subsection{Research Gaps}

Our analysis of the existing literature reveals several critical gaps:

\begin{itemize}
    \item \textbf{Lack of Automated 3D Pipeline}: No existing framework provides a fully automated pipeline for generating 3D road networks from publicly available GIS and elevation data.
    
    \item \textbf{Limited Elevation Integration}: Current co-simulation frameworks either ignore elevation entirely or require manual 3D environment creation.
    
    \item \textbf{Incomplete Validation}: Few studies provide comprehensive validation of their 3D environments against real-world topographical data.

    \item \textbf{Scalability Limitations}: Most existing approaches are demonstrated on small, manually created environments rather than real-world scale urban areas.
\end{itemize}

Our proposed framework addresses these gaps by providing an automated, scalable approach to elevation-aware 3D traffic co-simulation that can handle any real-world location with complex topographical features.

\section{Methodology}
\label{sec:methodology}

\begin{figure*}[t]
    \centering
    \includegraphics[width=0.85\textwidth]{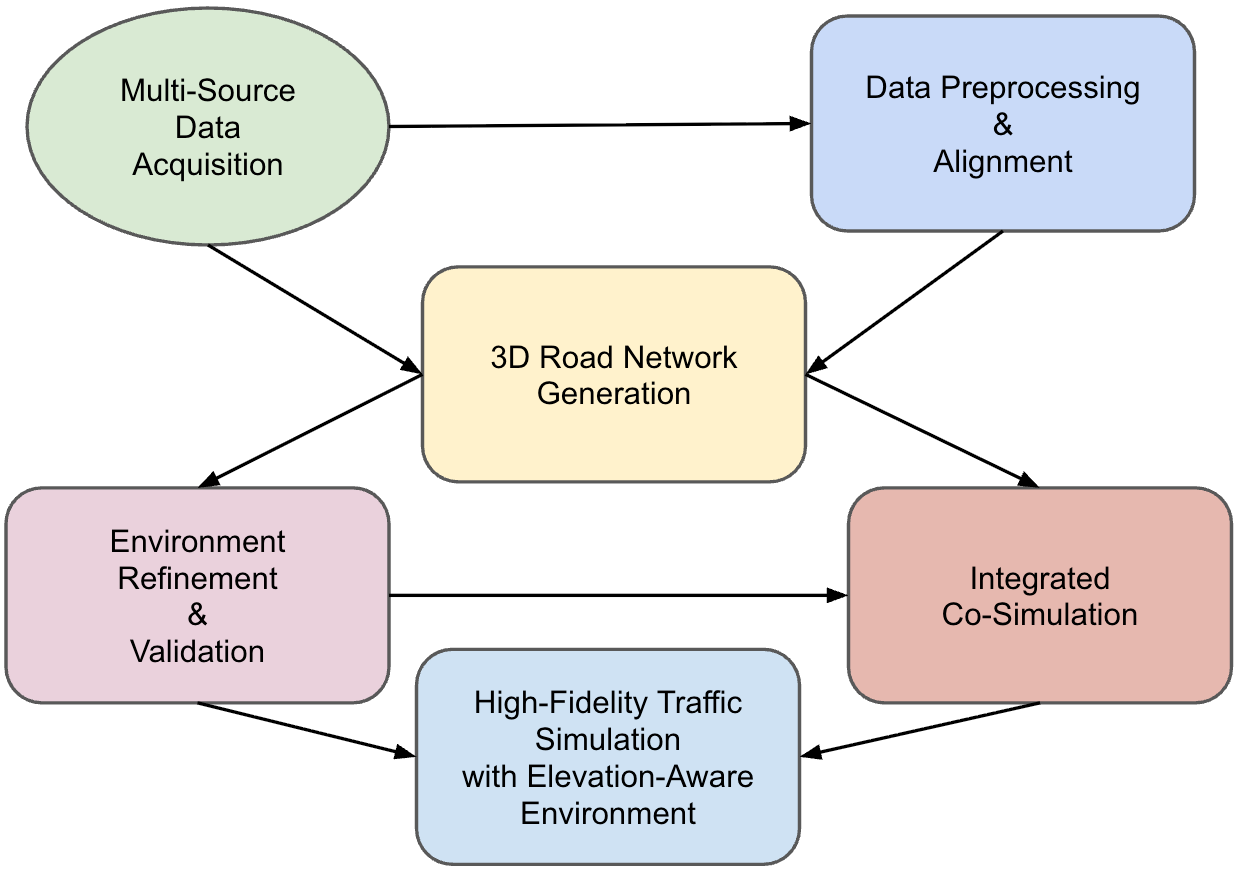}
    \caption{Overview of the elevation-aware 3D traffic co-simulation framework. The pipeline integrates multi-source geospatial data (OpenStreetMap and USGS DEM) through automated processing and 3D generation in RoadRunner, followed by validation and integrated co-simulation in CARLA-SUMO.}
    \label{fig:methodology_overview}
\end{figure*}

We presents an elevation-aware 3D traffic co-simulation framework that integrates 2D road network data with elevation information to create 3D representations for synchronized co-simulation between microscopic traffic modeling (SUMO) and high-fidelity 3D traffic simulator (CARLA). The framework addresses the critical need for elevation-aware simulation environments, as traditional flat-world assumptions fail to capture real-world driving dynamics on sloped terrain, which significantly impacts vehicle behavior, sensor perception, and autonomous driving algorithm performance. Figure~\ref{fig:methodology_overview} illustrates the complete pipeline architecture.

\subsection{Problem Formulation} 

The fundamental challenge lies in transforming readily available 2D road network data into accurate 3D representations that preserve real-world elevation profiles. Let $\mathcal{R}_{2D}$ denote a 2D road network extracted from OpenStreetMap, where each point is represented as $(x_i, y_i)$. Let $\mathcal{E}$ denote the elevation data from Digital Elevation Model (DEM), mapping 2D coordinates to height values. The objective is to construct a 3D road network $\mathcal{R}_{3D}$ by combining these datasets while ensuring geometric consistency, physical realism, and computational efficiency.

The following notation is used throughout:
\begin{itemize}
    \item $N$ = total number of road points
    \item $(x_i, y_i)$ = 2D coordinates from OpenStreetMap in UTM projection
    \item $z_i$ = elevation value at location $(x_i, y_i)$ in meters above sea level
    \item $(x_i, y_i, z_i)$ = 3D coordinate in the final road network
\end{itemize}

\subsection{Automated OSM-DEM Integration}
\label{sec:osm_dem_integration}

\subsubsection{Data Integration Process}

The integration of OpenStreetMap data and elevation data represents a critical step in creating realistic 3D environments. Traditional approaches often require manual elevation assignment or rely on flat approximations, leading to unrealistic road networks. The proposed automated pipeline eliminates these limitations through a systematic stacking operation. Given:
\begin{itemize}
    \item Road network points: $\mathcal{R}_{2D} = \{(x_1, y_1), (x_2, y_2), ..., (x_N, y_N)\}$
    \item Elevation data: $\mathcal{E} = \{z_1, z_2, ..., z_N\}$ at corresponding locations
\end{itemize}

The 3D road network is constructed as:
\begin{equation}
\mathcal{R}_{3D} = \mathcal{R}_{2D} \oplus \mathcal{E} = \{(x_i, y_i, z_i) \mid i = 1, 2, ..., N\}
\end{equation}
where $\oplus$ represents the stacking operation that appends elevation to each 2D coordinate. This operation is computationally efficient and preserves the topological structure of the original road network while adding the critical third dimension.

\subsubsection{Elevation Assignment}

Accurate elevation assignment is essential for realistic terrain representation. For each road point $(x_i, y_i)$ in the network, elevation is assigned using lookup and interpolation:
\begin{equation}
z_i = \mathcal{E}(x_i, y_i) = \text{Interpolate}(\text{DEM}, x_i, y_i)
\end{equation}

Since DEM data is typically provided as a regular grid with finite resolution (e.g., 1-meter spacing), direct lookup may not align with road coordinates. Therefore, bilinear interpolation is employed using the four nearest DEM grid points:
\begin{equation}
z_i = \sum_{j=1}^{4} w_j \cdot z_j^{\text{grid}}
\end{equation}
where $z_j^{\text{grid}}$ are the elevations at nearby grid points and $w_j$ are distance-based weights calculated as:
\begin{equation}
w_j = \frac{1/d_j}{\sum_{k=1}^{4} 1/d_k}
\end{equation}
with $d_j$ being the Euclidean distance from $(x_i, y_i)$ to grid point $j$. This weighted approach ensures smooth elevation transitions and prevents artifacts at grid boundaries.

The automated integration follows Algorithm 1:
\vspace{-1em}
\begin{algorithm}[H]
\caption{Automated OSM-DEM Integration}
\begin{algorithmic}[1]
\State \textbf{Input:} 2D road network $\mathcal{R}_{2D}$, DEM elevation data $\mathcal{E}$
\State \textbf{Output:} 3D road network $\mathcal{R}_{3D}$
\State Initialize empty $\mathcal{R}_{3D}$
\For{each point $(x_i, y_i)$ in $\mathcal{R}_{2D}$}
    \State $z_i \gets \mathcal{E}(x_i, y_i)$ \Comment{Look up elevation}
    \State Add $(x_i, y_i, z_i)$ to $\mathcal{R}_{3D}$
\EndFor
\For{each road segment}
    \State Calculate segment length $L$
    \State $n \gets \lfloor L / 1.0 \rfloor$ \Comment{Sample every meter}
    \For{$k = 1$ to $n-1$}
        \State - Interpolate position along segment
        \State - Assign elevation using DEM lookup
    \EndFor
\EndFor
\State \textbf{return} $\mathcal{R}_{3D}$
\end{algorithmic}
\end{algorithm}

\subsection{Elevation-Aware Coordinate Representation}
\label{sec:elevation_coords}

\subsubsection{3D Coordinate System} 

The choice of coordinate representation significantly impacts simulation accuracy and computational efficiency. Each point in the elevation-aware system is represented as:
\begin{equation}
P_i = (x_i, y_i, z_i) \in \mathbb{R}^3
\end{equation}
where:
\begin{itemize}
    \item $x_i$ = easting coordinate (meters) in UTM projection
    \item $y_i$ = northing coordinate (meters) in UTM projection
    \item $z_i$ = elevation above sea level (meters) from WGS84 datum
\end{itemize}

This representation maintains full 3D information without requiring complex transformations while ensuring compatibility with both SUMO (which traditionally operates in 2D) and CARLA (which requires full 3D coordinates).

\subsubsection{Gradient Analysis}

Road gradient directly affects vehicle dynamics, fuel consumption, and safety. Excessive gradients can make roads impassable for certain vehicle types and create dangerous driving conditions. For each road segment, the gradient (slope) is calculated to ensure physical realism:
\begin{equation}
\text{Gradient} = \frac{\Delta z}{\Delta d} = \frac{z_{\text{end}} - z_{\text{start}}}{\sqrt{(x_{\text{end}} - x_{\text{start}})^2 + (y_{\text{end}} - y_{\text{start}})^2}}
\end{equation}

Maximum gradient constraints are enforced based on road design standards and vehicle capabilities:
\begin{align}
|\text{Gradient}| &\leq \text{Max\_Gradient} \\
\shortintertext{where Max\_Gradient is defined as:}
\text{Max\_Gradient} &= 
\begin{cases}
    0.08 & \text{for highways (8\% grade)} \\
    0.12 & \text{for arterial roads (12\% grade)} \\
    0.15 & \text{for residential streets (15\% grade)}
\end{cases}
\end{align}

These thresholds are derived from civil engineering standards and ensure that generated roads remain navigable by standard vehicles.

\subsubsection{Elevation Profile Smoothing}

Raw elevation data may contain noise or abrupt changes that create unrealistic road profiles. When gradients exceed limits or show excessive variation, smoothing is applied to create drivable road surfaces:
\begin{equation}
z_i^{\text{smoothed}} = \frac{z_{i-1} + z_i + z_{i+1}}{3}
\end{equation}

This three-point moving average preserves the general terrain characteristics while eliminating local anomalies. The smoothing process is applied iteratively until gradient constraints are satisfied, ensuring both realism and drivability.

\begin{figure*}[t!]
    \centering
    \includegraphics[width=1\textwidth]{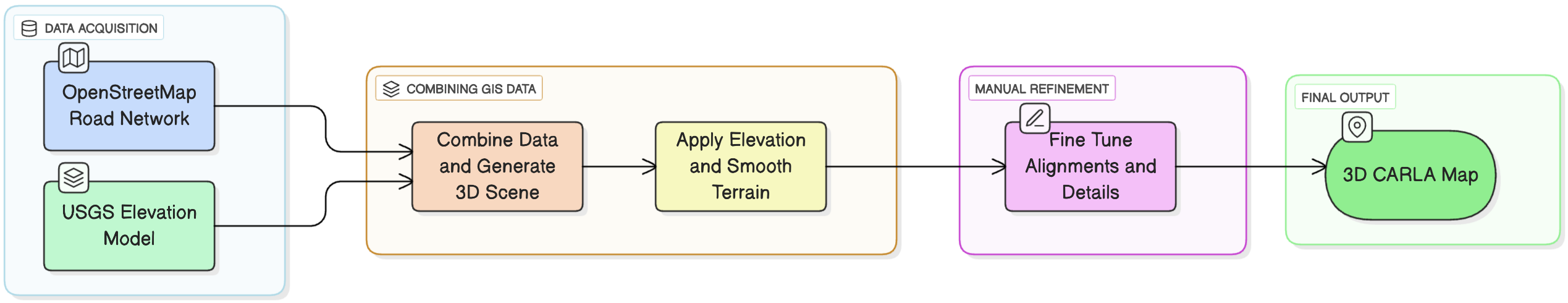}
    \caption{Creating 3D CARLA Maps by Integrating OSM and DEM Data in RoadRunner}
    \label{fig:RoadrunnerProcess}
\end{figure*}

\subsection{Validation Framework} 
\label{sec:validation}

\subsubsection{Elevation Accuracy Metrics}

Validating the accuracy of generated 3D road networks is essential for ensuring simulation fidelity. Multiple error metrics are employed to comprehensively assess elevation accuracy. The Mean Absolute Error (MAE) provides an average measure of elevation discrepancy:
\begin{equation}
\text{MAE} = \frac{1}{N} \sum_{i=1}^{N} |z_i^{\text{generated}} - z_i^{\text{actual}}|
\end{equation}

For a more statistically robust measure that penalizes larger errors, the Root Mean Square Error (RMSE) is calculated:
\begin{equation}
\text{RMSE} = \sqrt{\frac{1}{N} \sum_{i=1}^{N} (z_i^{\text{generated}} - z_i^{\text{actual}})^2}
\end{equation}

Additionally, the maximum error is tracked to identify worst-case deviations:
\begin{equation}
\text{Max\_Error} = \max_{i} |z_i^{\text{generated}} - z_i^{\text{actual}}|
\end{equation}

These metrics are computed against high-resolution ground truth data when available, or against the original DEM data to assess interpolation accuracy. For comprehensive spatial analysis, the 3D Euclidean distance error is also calculated:
\begin{equation}
\text{Error}_{3D} = \sqrt{(x_i^{\text{gen}} - x_i^{\text{act}})^2 + (y_i^{\text{gen}} - y_i^{\text{act}})^2 + (z_i^{\text{gen}} - z_i^{\text{act}})^2}
\end{equation}

\subsubsection{Geometric Consistency Verification}

Beyond absolute accuracy, the geometric consistency of the road network must be verified to ensure realistic driving conditions. Road gradients are checked against design constraints:

\begin{equation}
\text{Gradient\_Compliance} = \frac{\text{Number\ of\ Valid\ Segments}}{\text{Total\ Segments}} \times 100\%
\end{equation}

A segment is considered valid if its gradient satisfies the constraints defined in Equation (7). This metric ensures that the generated roads remain within drivable limits throughout the network.

\subsubsection{Network Connectivity Assessment}

At intersections, elevation continuity is crucial for smooth vehicle transitions between road segments. Discontinuities can cause simulation artifacts and unrealistic vehicle behavior. Elevation gaps at connections are constrained:
\begin{equation}
\text{Elevation\_Gap} = |z_{\text{road1}} - z_{\text{road2}}| < 0.1 \text{ meters}
\end{equation}

This threshold of 0.1 meters prevents noticeable jumps while allowing for minor variations due to different road construction standards or data resolution limits.

\subsection{Co-Simulation Integration}
\label{sec:cosimulation}

\subsubsection{Temporal Synchronization}

Accurate co-simulation requires precise temporal synchronization between SUMO's microscopic traffic simulation and CARLA's 3D physics engine. Both simulators advance in lockstep with a fixed time step:
\begin{equation}
t_{\text{SUMO}} = t_{\text{CARLA}} = n \times \Delta t
\end{equation}
where $\Delta t = 0.05$ seconds (20 Hz update rate) and $n$ is the step counter. This update rate balances simulation accuracy with computational efficiency, providing sufficient temporal resolution for realistic vehicle dynamics while maintaining real-time performance.

\subsubsection{State Exchange Protocol}

At each time step, comprehensive vehicle states are exchanged between simulators to maintain consistency:
\begin{equation}
\text{Vehicle\_State} = (x, y, z, \text{speed}, \text{heading})
\end{equation}

The elevation $z$ for each vehicle is crucial for accurate 3D positioning and is determined by interpolating the vehicle's position on the road network:
\begin{equation}
z_{\text{vehicle}} = \mathcal{R}_{3D}(x_{\text{vehicle}}, y_{\text{vehicle}})
\end{equation}

This ensures that vehicles follow the terrain accurately, maintaining contact with the road surface and exhibiting realistic physics behavior on slopes.

\subsubsection{Synchronization Error Management}

Due to differences in internal representations and numerical precision, position discrepancies may accumulate between simulators. It is important to note that CARLA's synchronization interface operates primarily in 2D coordinates, as the elevation is implicitly determined by the road surface. Therefore, synchronization error monitoring focuses on horizontal position differences:
\begin{equation}
\text{Sync\_Error} = \sqrt{(x_{\text{SUMO}} - x_{\text{CARLA}})^2 + (y_{\text{SUMO}} - y_{\text{CARLA}})^2}
\end{equation}

The elevation component is not included in this error calculation because CARLA automatically adjusts vehicle height based on the underlying terrain. When Sync\_Error exceeds 0.5 meters, resynchronization is triggered to maintain consistency. This threshold balances accuracy requirements with computational overhead, preventing excessive resynchronization while ensuring vehicles remain properly positioned.

\begin{figure*}[t!]
    \centering
    \includegraphics[width=0.9\textwidth]{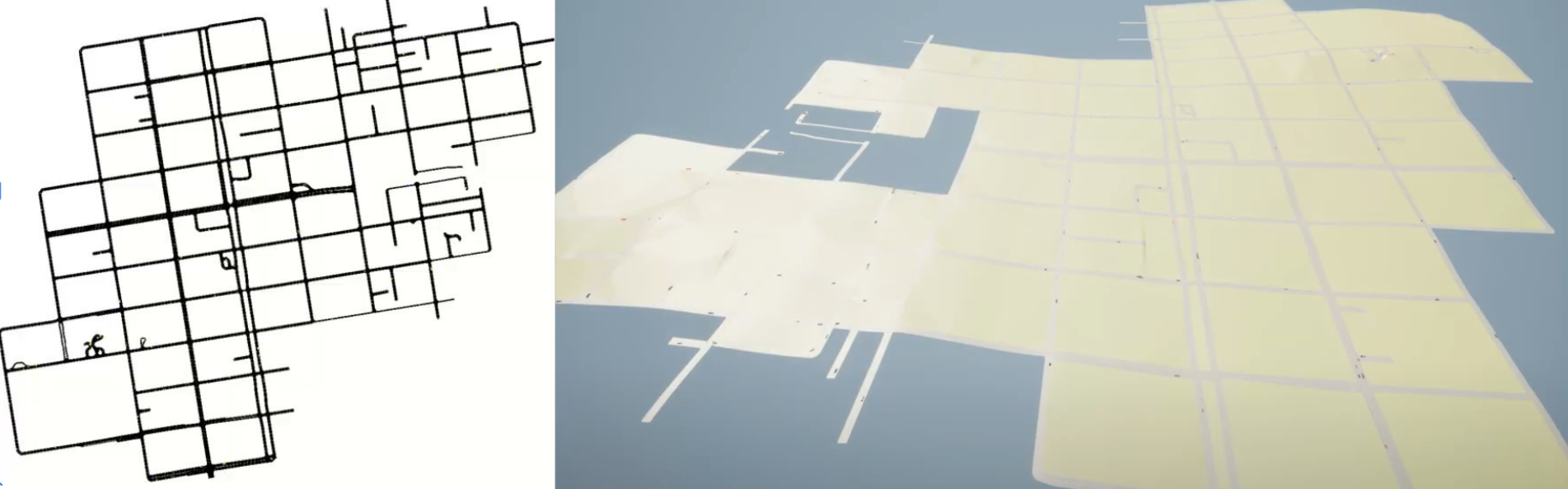}
    \caption{Side-by-Side Visualization of a Synchronized Co-Simulation, Correlating the 2D Microscopic Traffic View in SUMO with the 3D High-Fidelity Vehicle View in CARLA}
    \label{fig:co_simulation_result_NE}
\end{figure*}

\subsection{Real-World Scalability}
\label{sec:scalability} 

\subsubsection{Large-Scale Urban Network Generation}

Unlike previous approaches that are limited to small, manually created environments, this framework enables the creation of elevation-aware 3D road networks for real-world urban areas at scale. For instance, the methodology has been successfully applied to different areas of San Francisco, generating 3D road networks with hundreds of intersections and road segments that accurately reflect real-world topography.

The framework processes real OpenStreetMap data directly, extracting complex road networks with multiple lane configurations, traffic signals, and intersection geometries that represent actual urban infrastructure. Each road segment incorporates precise elevation data from USGS DEM sources, creating realistic driving conditions that include hills, valleys, and varying terrain profiles typical of metropolitan areas.

\subsubsection{Real-World Traffic Demand Integration}

A key advantage over existing approaches is the ability to integrate real-world traffic demand patterns with elevation-aware 3D simulation. 
This methodology enables:

\begin{itemize}
    \item Direct integration of real traffic demand data into SUMO's microscopic traffic simulation
    \item Elevation-aware vehicle routing that accounts for road gradients and terrain effects on traffic flow
    \item Realistic traffic patterns that reflect actual urban mobility patterns
    \item Multi-modal traffic simulation including cars, trucks, buses, and other vehicle types with different elevation handling characteristics
\end{itemize}

\subsubsection{Scalability Across Multiple Dimensions}

The framework demonstrates scalability across multiple dimensions:

\begin{itemize}
    \item \textbf{Road Network Size}: Successfully processes metropolitan areas with hundreds of intersections and road segments, compared to previous works limited to small, artificial road networks without elevations
    \item \textbf{Traffic Volume}: Handles high-fidelity simulations with hundreds or thousands of vehicles, enabling realistic congestion scenarios in 3D environments
    \item \textbf{Computational Efficiency}: Processes large networks in practical timeframes; for example, a road network with $\approx 100$ intersections is processed in under 8 minutes on standard hardware ({i9-13900K / RTX 4080}).
\end{itemize}

This scalability enables realistic autonomous vehicle testing scenarios that previous approaches cannot support, bridging the gap between synthetic simulation environments and real-world deployment requirements.

\subsection{Performance Optimization}

To handle large-scale simulations with hundreds or thousands of vehicles efficiently, optimization strategies are essential for maintaining real-time performance.

\subsubsection{Distance-Based Rendering}

CARLA leverages Unreal Engine's built-in level-of-detail (LOD) system to automatically optimize rendering performance. Custom LOD configurations can be defined through Unreal Engine's asset pipeline if specific optimization requirements arise. This automatic optimization significantly reduces computational load while maintaining visual quality near the ego vehicle, enabling larger-scale simulations without sacrificing fidelity in the immediate vicinity.
\newline

\noindent This methodology provides a comprehensive framework for elevation-aware 3D traffic co-simulation that addresses the limitations of traditional flat-world approaches through:

\begin{itemize}
    \item \textbf{Gap 1 (Automated 3D Pipeline):} Systematic stacking operation $\mathcal{R}_{3D} = \mathcal{R}_{2D} \oplus \mathcal{E}$ that automates the integration of 2D road networks with elevation data, eliminating manual 3D modeling efforts

    \item \textbf{Gap 2 (Elevation Integration):} Direct 3D coordinate representation $(x, y, z)$ with gradient constraints and smoothing ensuring physical realism and drivability across diverse terrains

    \item \textbf{Gap 4 (Validation):} Comprehensive validation framework with multiple error metrics (MAE, RMSE, Max Error) and geometric consistency checks ensuring accurate elevation representation in complex urban environments

    \item \textbf{Gap 3 (Scalability):} Real-world applicability to metropolitan areas like San Francisco with hundreds of intersections and road segments, supporting both large road networks and high-volume traffic scenarios that previous approaches cannot handle
\end{itemize}

This integrated approach enables realistic autonomous vehicle testing in complex urban environments with accurate elevation profiles, providing a critical tool for developing and validating algorithms that must operate in real-world topographies.
\section{Experiments}
\label{sec:experiments}

\subsection{Elevation Aware Map Generation}
\label{Exp:Map Generation}
\subsubsection{Target Regions}

San Francisco was selected as our primary test environment due to its unique topographical challenges that stress-test our framework's capabilities. We tested four different parts of the city to comprehensively evaluate different type of elevation scenarios:

\begin{itemize}
    \item \textbf{Northeast Quadrant}: Features the iconic hills of Russian Hill and Nob Hill, with maximum gradients exceeding 30\% and complex intersection geometries at elevation transitions.
    \item \textbf{Southeast quadrant}: Arterials and waterfront connectors, including segments with gentle slopes and elevated structures.
    \item \textbf{Northwest quadrant}: Residential grid with rolling terrain and park-edge elevation changes.
    \item \textbf{Southwest quadrant}: Freeway corridors and interchanges featuring sustained grades and merges.
\end{itemize}

\begin{figure}[t]
    \centering
    \includegraphics[width=0.48\textwidth]{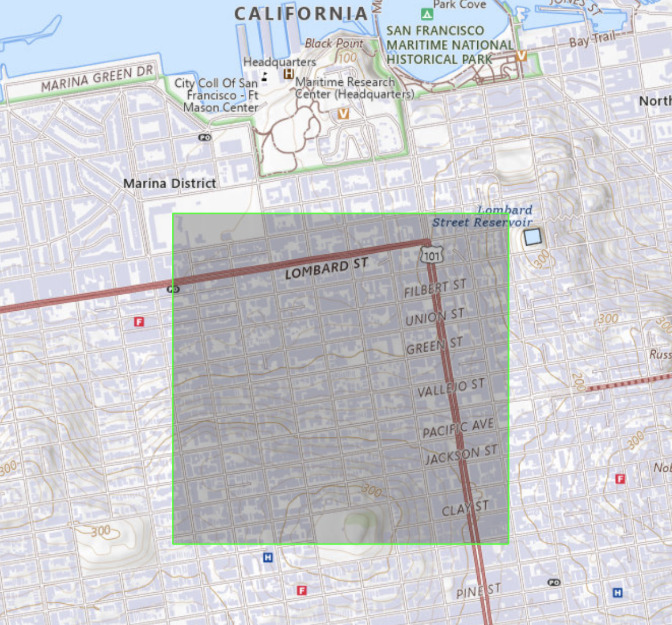}
    \caption{Example map extent for the San Francisco North-East quadrant used in Objective 1.}
    \label{fig:SF-NE-Area}
\end{figure}

Each quadrant spans a compact urban area with a dense road network and multiple intersections, providing diverse testing scenarios for our framework.

\subsubsection{Data Sources}

Each quadrant uses OpenStreetMap (OSM) for the 2D road network and USGS Digital Elevation Models (DEM, 1/3 arc-second) for terrain height. Geographic coordinates are projected from WGS84 to UTM and converted to a local Cartesian frame for simulation consistency.

\subsubsection{Processing Pipeline}

Map generation follows the automated OSM--DEM integration formalized in Section~\ref{sec:osm_dem_integration} and the elevation-aware coordinate representation in Section~\ref{sec:elevation_coords}:
\begin{enumerate}
    \item \textbf{Network extraction}: OSM ways and intersections are parsed within a quadrant bounding box; attributes and lane topology are retained.
    \item \textbf{Elevation assignment}: Per-vertex heights are from DEM; captures terrain variation at the specified region.
    \item \textbf{Gradient constraints \/smoothing}: Elevation profiles are adjusted to enforce physically reasonable grades while preserving DEM fidelity.
    \item \textbf{Export and validation hooks}: The resulting 3D network is exported to OpenDRIVE and 3D Assets for CARLA and to \texttt{.net.xml} for SUMO, enabling cross-simulator consistency checks as in Section~\ref{sec:cosimulation}.
\end{enumerate}

\subsection{Planned Experiments Using the Generated Maps}

The maps produced in \ref{Exp:Map Generation} will be used in the following planned studies; quantitative analyses will be reported in later sections once experiments complete:
\begin{itemize}
    \item \textbf{Elevation accuracy and geometric consistency} (Section~\ref{sec:validation}): Evaluate MAE/RMSE against reference elevation sources, gradient compliance, and intersection continuity.
\begin{equation}
\scalebox{0.9}{$
RMSE_{elevation} = \sqrt{\frac{1}{n} \sum_{i=1}^{n} (E_{generated,i} - E_{ground\_truth,i})^2}
$}
\end{equation}

    \item \textbf{Co-simulation synchronization} (Section~\ref{sec:cosimulation}): Measure position-consistency bounds and resynchronization behavior across SUMO--CARLA.
    \item \textbf{Scalability and throughput} (Section~\ref{sec:scalability}): Benchmark preprocessing/runtime vs. network size, sampling density, and parallelism.
    \item \textbf{Downstream tasks}: Qualitatively assess perception and planning behaviors on sloped terrain using the generated 3D maps.
\end{itemize}

\section{Results}

This section presents preliminary qualitative results for the elevation-aware 3D maps for SUMO--CARLA co-simulation. Quantitative analyses are yet to be added until final experiments are complete.

\subsection{Preliminary Qualitative Outcomes}

\textbf{Successful 3D map generation:} Using the automated OSM--DEM pipeline (Section~\ref{sec:osm_dem_integration}), elevation-aware 3D urban road networks are produced. The resulting maps preserve topological structure while incorporating terrain height throughout intersections, segments, and ramps.

\textbf{Continuous elevation profiles and smooth gradients:} Visual inspection confirms that roadway elevation varies smoothly along segments with appropriate transitions at intersections, aligning with the gradient constraints described in Section~\ref{sec:elevation_coords}. Steep urban streets, rolling residential terrain, and elevated connectors are represented with consistent slopes and continuous geometry.

\textbf{Co-simulation readiness:} The generated maps with consistent topology and elevation, allowed synchronized 2D--3D co-simulation as per Section~\ref{sec:cosimulation}. The rendered CARLA scenes exhibit plausible slopes and drivable surfaces, and vehicle trajectories in SUMO project coherently onto 3D road surfaces.

\begin{figure}[t]
    \centering
    \includegraphics[width=0.48\textwidth]{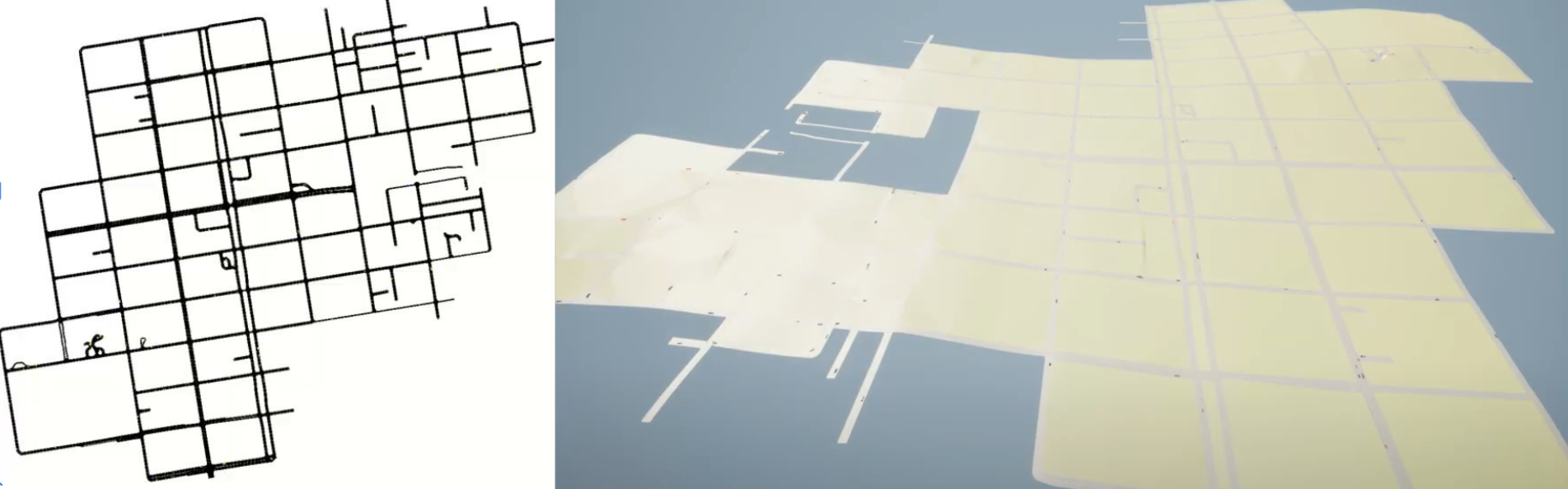}
    \caption{Example co-simulation visualization for the North-East quadrant: 2D microscopic traffic state (SUMO) is synchronized with 3D vehicle motion (CARLA) on elevation-aware roads.}
    \label{fig:qual_co_sim_ne}
\end{figure}

\begin{figure}[t]
    \centering
    \includegraphics[width=0.48\textwidth]{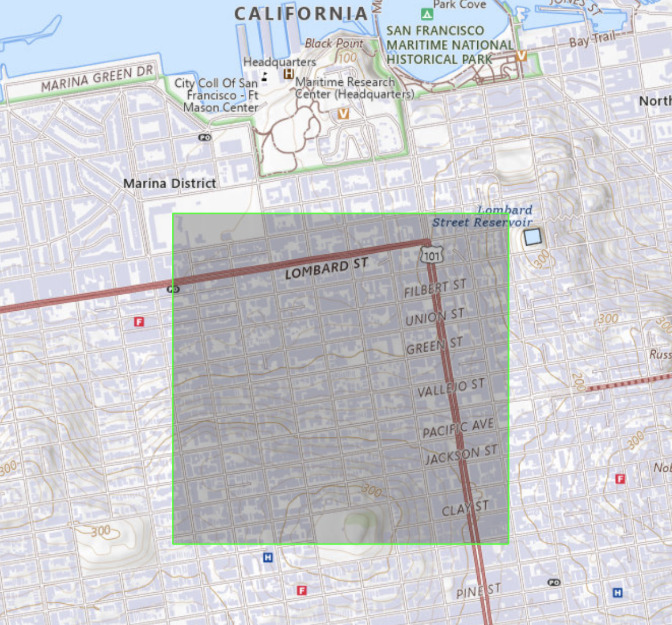}
    \caption{Rendered map extent used to generate an elevation-aware 3D scene for the North-East quadrant.}
    \label{fig:extent_ne}
\end{figure}

\subsection{Scope and Pending Analyses}

These results confirm that the map generation pipeline produces co-simulation ready digital twin environment with coherent elevation structure across diverse road networks. Planned analyses will report quantitative accuracy, synchronization behavior, and scalability once experiments are finalized.

\section{Conclusion}
This work introduces a unified, elevation-aware 3D co-simulation framework that bridges the capabilities of SUMO’s large-scale microscopic traffic models and CARLA’s high-fidelity perception and vehicle dynamics engine. By automating the fusion of OSM road networks with DEM elevation data, the proposed pipeline produces realistic 3D environments that reflect real-world topography—addressing a long-standing limitation of existing simulation tools. Our methodology ensures geometric consistency, enforces physical road design constraints, and enables seamless deployment of the resulting digital twin environments into synchronized SUMO–CARLA simulations. Qualitative results across multiple topographically diverse regions of San Francisco demonstrate the framework’s ability to model steep grades, rolling hills, and complex intersection geometries while preserving drivability and simulation coherence.

Although quantitative evaluations are ongoing, the current outcomes confirm the pipeline’s robustness and scalability, supporting real-world urban networks with hundreds of intersections and enabling perception-centric AV experiments under realistic elevation conditions. This work lays the foundation for future extensions, including broader geographic deployment, enhanced sensor modeling, and integration of real traffic demand. By capturing elevation as a first-class component of simulation, the framework moves the field closer to deployment-ready digital twins for autonomous vehicle development. We plan to further test the framework within mixed traffic control scenarios~\cite{Pan2025Review,Liu2025Large,Islam2025Heterogeneous,Fan2025OD,Wang2024Intersection,Wang2024Privacy,Poudel2024CARL,Poudel2024EnduRL,Villarreal2024Eco,Villarreal2023Pixel,Villarreal2023Chat}.

{
    \small
    \bibliographystyle{ieeenat_fullname}
    \bibliography{main}
}


\end{document}